\documentclass[conference]{IEEEtran}
\IEEEoverridecommandlockouts

\usepackage{algorithm}
\usepackage{amsthm}              

\usepackage{booktabs,makecell} 
\usepackage{graphicx}     
\graphicspath{{figs/}}     

\usepackage{svg}  
\svgpath{{figs/}} 
\svgsetup{inkscapelatex=false}
\usepackage{hyperref}

\usepackage{url} 
\usepackage{float}
\usepackage{cite}
\usepackage{amsmath,amssymb,amsfonts}
\usepackage{algorithmic}
\usepackage{graphicx}
\usepackage{textcomp}
\usepackage{xcolor}
\def\BibTeX{{\rm B\kern-.05em{\sc i\kern-.025em b}\kern-.08em
    T\kern-.1667em\lower.7ex\hbox{E}\kern-.125emX}}
\begin{document}

\title{HATS: High-Accuracy Triple-Set Watermarking for Large Language Models
\thanks{This research was supported by the National Natural Science Foundation of China (Grant No.  62302468, U2441239).\\
Correspondence:luxaole@gmail.com}
}
\author{
\IEEEauthorblockN{
1\textsuperscript{st} Zhiqing Hu\IEEEauthorrefmark{4}\IEEEauthorrefmark{2}, 
2\textsuperscript{nd} Chenxu Zhao\IEEEauthorrefmark{4}\IEEEauthorrefmark{2}, 
3\textsuperscript{rd} Jiazhong Lu\IEEEauthorrefmark{3}, 
4\textsuperscript{th} Xiaolei Liu\IEEEauthorrefmark{2}\IEEEauthorrefmark{4}\textsuperscript{*}
}
\IEEEauthorblockA{\IEEEauthorrefmark{4}Institute of Computer Application, China Academy of Engineering Physics, Mianyang, China}
\IEEEauthorblockA{\IEEEauthorrefmark{2}National Interdisciplinary Research Center of Engineering Physics, Mianyang, China}
\IEEEauthorblockA{\IEEEauthorrefmark{3}School of Cybersecurity (Xin Gu Industrial College), Chengdu University of Information Technology, Mianyang, China}
\IEEEauthorblockA{Email: huzhiqing23@gscaep.ac.cn, zhaochenxu25@gscaep.ac.cn, ljz@cuit.edu.cn, luxaole@gmail.com}
}

\maketitle 

\begin{abstract}
Misuse of LLM-generated text can be curbed by watermarking techniques that embed implicit signals into the output. We propose a watermark that partitions the vocabulary at each decoding step into three sets (Green/Yellow/Red) with fixed ratios and restricts sampling to the Green and Yellow sets. At detection time, we replay the same partitions, compute Green-enrichment and Red-depletion statistics, convert them to one-sided $z$-scores, and aggregate their $p$-values via Fisher's method to decide whether a passage is watermarked. We implement generation, detection, and testing on \texttt{Llama 2 7B}, and evaluate true-positive rate, false-positive rate, and text quality. Results show that the triple-partition scheme achieves high detection accuracy at fixed FPR while preserving readability.
\end{abstract}



\begin{figure*}[t]
  \centering
  \includegraphics[width=0.95\textwidth]{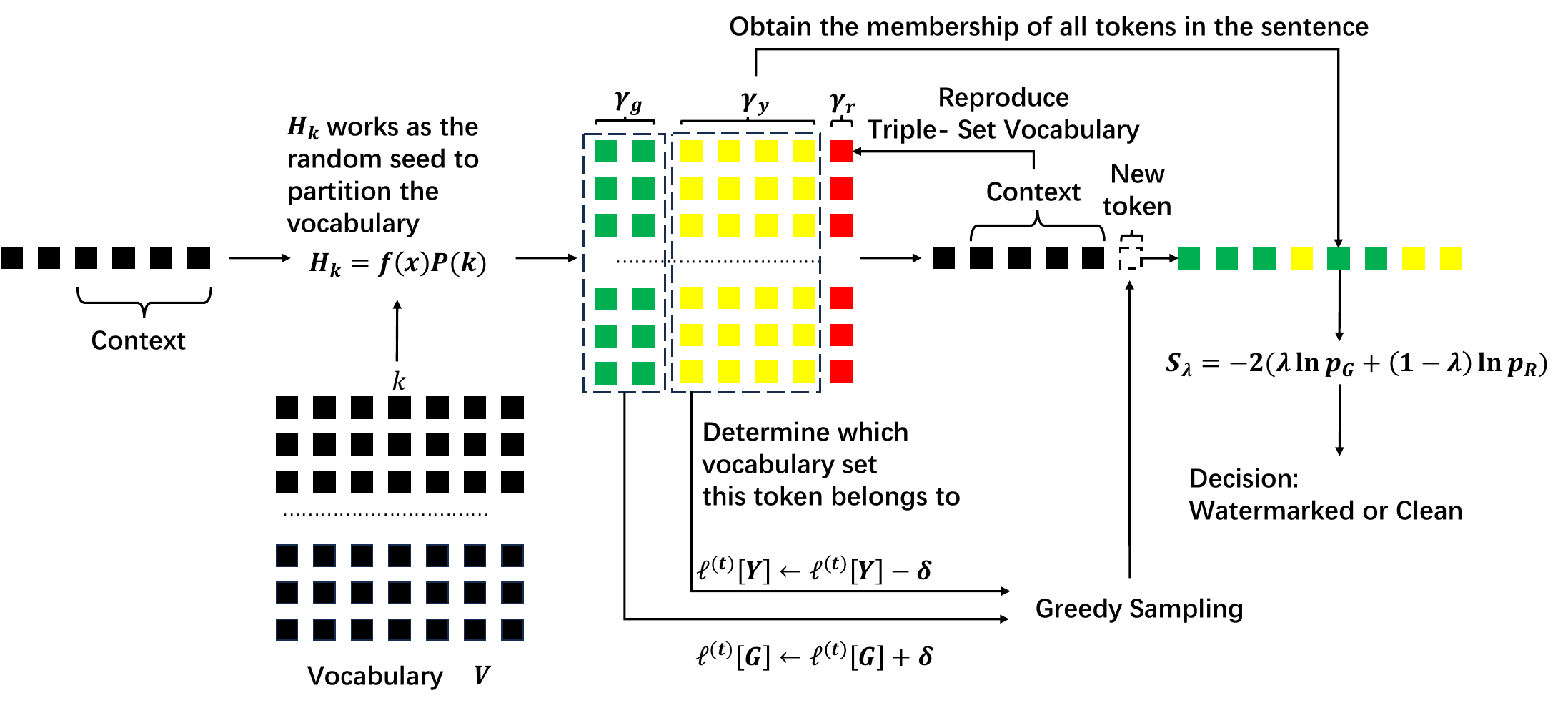}
  \caption{Overall framework of the proposed method.}
  \label{fig:longfig}
\end{figure*}

\section{Introduction}
In recent years, the natural language understanding and generation capabilities of Large Language Models (LLMs) have improved significantly, leading to their widespread adoption across various domains of production and daily life~\cite{b1,b2,b4}. 
With the emergence of dialogue-based models such as ChatGPT~\cite{b3}, LLaMA~\cite{b10}, and DeepSeek~\cite{b11}, general-purpose language models are rapidly being integrated into critical sectors including education, finance, healthcare, and public opinion analysis. 
However, the proliferation of such models has also introduced potential security and ethical risks. 
Their misuse for malicious purposes---including social engineering attacks and election manipulation via automated social bots, as well as the propagation of misinformation and synthetic online content---has become an increasing concern~\cite{b5,b6,b27}. 
These risks have sparked widespread attention in both academia and industry regarding the traceability and authenticity of generative content.

To ensure accountability and authenticity of model outputs without compromising semantic quality, watermarking techniques for LLMs have emerged as a promising solution. 
Large language model watermarking aims to embed identifiable yet imperceptible patterns into generated text, thereby supporting responsible AI usage, abuse prevention, and provenance verification.

The evolution of text watermarking is a long process. 
Early methods were format-based, embedding watermarks by modifying layout rather than content. 
For instance, Fung et al.~\cite{b12} encoded information by vertically or horizontally shifting text lines or words, while Por et al.~\cite{b13} embedded Unicode whitespace characters between words. 
However, such approaches were easily broken after text reformatting, resulting in poor robustness. 
Later, lexical substitution-based methods were proposed. 
Topkara et al.~\cite{b14} introduced synonym replacement using WordNet~\cite{b15} as the synonym source, and watermark detection was performed by applying inverse substitution rules. 
With the rise of deep learning, generation-based watermarking methods emerged. 
Liu and Bu~\cite{b8} developed AWT, an end-to-end scheme that encodes sentences and watermark messages jointly through Transformer-based architectures, enabling hidden information extraction at decoding time.

As LLMs became prevalent, watermarking during text generation gained significant attention due to its ability to preserve naturalness while embedding signals directly in the generative process. 
Kirchenbauer \emph{et al.}~\cite{b7} proposed \textbf{KGW}, which partitions the vocabulary into green and red lists and samples tokens from the green list using a secret key, enabling detection through green-token proportion analysis. Despite its simplicity, it yields moderate accuracy and variable text quality. 

Building on this line of research, our work focuses on enhancing watermark detection accuracy.
We propose \textbf{HATS}, a watermarking approach that introduces a three-color token distribution pattern within generated text.  
Compared with KGW~\cite{b7}, our method adds an additional watermark signal and refines the detection strategy, achieving higher true positive rates, lower false positive rates, while maintaining relatively high text quality.

\section{Related Work}
Existing watermarking methods for large language models can be broadly categorized into two classes: \textit{rewriting-based} and \textit{generation-based} approaches.

The first category modifies existing text by lexical or semantic rewriting. 
Yang et al.~\cite{b9} proposed a context-aware lexical substitution watermarking scheme that embeds identification signals by replacing selected words with contextually appropriate synonyms, guided by a secret key. 
During detection, the system compares the co-occurrence patterns and contextual consistency of replaced words to verify embedded information. 
This method is model-agnostic and suitable for external text provenance tracing, but it suffers from limited embedding capacity and sensitivity to human editing. 
To improve capacity, Yoo et al.~\cite{b17} designed a multi-bit natural language watermarking framework that maps multiple binary bits onto lexical or syntactic variants through probabilistic assignment. 
The method balances payload and linguistic naturalness by integrating semantic similarity filters and language model perplexity constraints.

The second category embeds watermarks during model inference. 
Kirchenbauer et al.~\cite{b7} introduced a token-level watermarking algorithm for LLMs that divides the vocabulary into ``green'' and ``red'' sets using a hash function. 
During generation, the model samples tokens preferentially from the green set according to the current context and key, producing a subtle statistical bias. 
Detection then relies on testing the proportion of green tokens within the generated text. 
While lightweight and effective, this method remains sensitive to paraphrasing and regeneration. 
To address these issues, Kirchenbauer et al.~\cite{b16} further investigated the reliability of LLM watermarks, analyzing their statistical stability under various decoding strategies (e.g., temperature sampling, top-$p$ truncation) and proposing sliding-window detection with adaptive thresholds to control false positives. 
Their study formally linked LLM watermarking to statistical significance testing, paving the way for verifiable detection.

Following this direction, Fernandez et al.~\cite{b19} proposed the ``Three Bricks'' framework, which unifies and standardizes watermark evaluation from a robustness perspective. 
The framework consists of three modules: (1) \textit{robust embedding}, enhancing signal strength through improved distributional bias and noise suppression; (2) \textit{multi-scale detection}, applying sliding-window and hierarchical statistical tests to adapt to varying text lengths; and (3) \textit{adversarial evaluation}, systematically testing watermark survival under paraphrasing, compression, and concatenation attacks. 
Although not a new algorithm, the framework provides a consistent benchmark for assessing watermark reliability and reproducibility.

In summary, while existing methods have demonstrated the feasibility of textual watermarking, most approaches face a low accuracy in detection stage. 
Our proposed HATS method extends prior work by introducing multi-level signal embedding and an enhanced statistical detection framework, contributing to more reliable provenance verification in large language model outputs.

\section{Methodology}

\subsection{Review of the KGW Method}
At time step $t$, the KGW method~\cite{b7} generates a random seed based on the context $\{x_{t-h}, \ldots, x_{t-2}, x_{t-1}\}$ to perform a stochastic partition of the vocabulary $V$, where the green list occupies $\gamma |V|$ entries and the red list occupies $(1-\gamma)|V|$ entries. To address low-entropy cases, the method adds a bias $\delta$ to the probability distribution over the green list. 

During the detection phase, the authors assume that the token-level statistics follow a Gaussian distribution.

Null hypothesis $H_0$:
The text is not watermarked (i.e., no bias is applied on $G_t$).  
Let the hit indicator at position $i$ be defined as
\begin{equation}
s_i^G = \mathbf{1}\{x_i \in G_i\}.
\end{equation}
Conditioned on the same key and partitioning scheme, we have
$\mathbb{E}[s_i^G] = \Pr(x_i \in G_i) = \gamma$, and approximately, for a window of length $L$,
\begin{equation}
S_G = \sum_{i=1}^{L} s_i^G \sim \mathrm{Binomial}(L, \gamma),
\qquad
\hat{p}_G = \frac{S_G}{L}.
\end{equation}
Thus, the one-sided $z$-statistic for enrichment is
\begin{equation}
z = \frac{\hat{p}_G - \gamma}{\sqrt{\gamma(1-\gamma)/L}}
\ \overset{H_0}{\approx}\ \mathcal{N}(0,1).
\end{equation}
And $H_0$ is rejected on the right tail (i.e., significant green enrichment).

The original two-list watermarking approach suffers from a relatively low detection success rate and insufficient robustness under random editing or deletion attacks.
To address these limitations, we enhance the baseline from two perspectives: the generation stage and the detection stage.
In the generation stage, the proposed HATS mechanism partitions the vocabulary into three disjoint subsets—red, yellow, and green—providing two complementary watermarking signals.
In particular, the red list embeds a strong watermark signal that significantly improves detection accuracy and robustness while preserving semantic naturalness.
In the detection stage, we introduce a new hypothesis model based on the Poisson–Binomial distribution, which stabilizes detection performance.

\subsection{Generation Stage}
During the generation stage, the vocabulary $V$ is partitioned into three disjoint subsets---green, yellow, and red. 
Tokens belonging to the \textit{green list} are assigned a positive bias in their probability distribution to increase the likelihood of being selected. 
Tokens in the \textit{yellow list} receive a slight negative bias to enhance semantic diversity under low-entropy conditions, 
whereas tokens in the \textit{red list} are strictly prohibited from selection, serving as a strong watermark signal embedded within the text. 
To prevent noticeable semantic distortion, the proportion of tokens assigned to the red list should be kept relatively small.

\begin{algorithm}[ht]
\caption{Generation Stage}
\label{alg:tri-embed-nogate-nokey}
\begin{algorithmic}[1]
\REQUIRE prompt $x_{1:N_p}$;\; ratios $(\gamma_g,\gamma_y,\gamma_r)$;\; bias amplitude $\delta>0$
\ENSURE completion $x_{N_p+1:N}$
\STATE \textbf{Constants:} seeding$=\textsc{lefthash}$,\; window $h{=}4$
\FOR{$t \gets N_p{+}1$ \TO $N$}
  \STATE Obtain logits $\ell^{(t)}$ from context $x_{1:t-1}$
  \STATE $c_t \leftarrow (x_{\max(1,t-h)},\ldots,x_{t-1})$
  \STATE $u \leftarrow \textsc{PRF}\!\big(\textsc{HASH}(c_t)\big)$ \hfill{\small}
  \STATE $G\gets\{k:\ u^{(G)}_k<\gamma_g\}$;\quad
         $R\gets\{k\in\overline{G}:\ u^{(YR)}_k<\tfrac{\gamma_r}{1-\gamma_g}\}$;\quad
         $Y\gets \overline{G}\setminus R$
  \STATE \textbf{Biasing:}\quad
         $\ell^{(t)}[G]\!\leftarrow\!\ell^{(t)}[G]+\delta$;\quad
         $\ell^{(t)}[Y]\!\leftarrow\!\ell^{(t)}[Y]-\delta$;\quad
         $\ell^{(t)}[R]\!\leftarrow\!-\infty$
  \STATE \textbf{Greedy selection:}\quad
         $x_t \leftarrow \arg\max_k\, \ell^{(t)}_k$
\ENDFOR
\end{algorithmic}
\end{algorithm}

\paragraph*{Seeding Scheme}
The term \textit{seeding} in the algorithm refers to which part of the context is included in the hash computation. The \textit{LeftHash} strategy with $h=4$ means that, at time step $t$:
\begin{itemize}
  \item Only the most recent four generated tokens are considered: $x_{t-4},\, x_{t-3},\, x_{t-2},\, x_{t-1}$.
  \item To form a random seed, each of these four tokens' IDs is passed through a fixed hash function $H(\cdot)$, and then combined using a \textbf{commutative aggregation} operation (e.g., bitwise XOR $\oplus$ or summation modulo $2^{64}$) to produce a 64-bit seed.
\end{itemize}

The three-way partitioning of the vocabulary is pseudorandom, ensuring reproducibility of the word lists during the detection phase.

\subsection{Detection Stage}
\paragraph*{Poisson–binomial in detection}
At detection time we count, within each window, how many positions hit a target set (Green or Red). A simple view is that these hits are Bernoulli trials whose success probabilities need not be identical across positions, because they change with context, special-token exclusion, sampling truncation (e.g., top-$k$/nucleus), and the evolving partitions induced by the seeding scheme. Modeling the window count with a \emph{Poisson–binomial} distribution therefore matches the data-generating process better than a single-parameter binomial.

\begin{algorithm}[ht]
\caption{Detection Stage}
\label{alg:tri-detect-nowindow-nokey}
\begin{algorithmic}[1]
\REQUIRE text $x_{1:N}$;\; ratios $(\gamma_g,\gamma_r)$;\; Fisher weight $\lambda_f$;\; global level $\alpha$
\ENSURE decision $\in\{\textsc{Watermarked},\textsc{Clean}\}$
\STATE \textbf{Constants:} seeding$=\textsc{lefthash}$,\; window $h{=}4$
\FOR{$i \gets 1$ \TO $N$}
  \STATE $c_i \leftarrow (x_{\max(1,i-h)},\ldots,x_{i-1})$
  \STATE $u \leftarrow \textsc{PRF}\!\big(\textsc{HASH}(c_i)\big)$ \hfill{\small }
  \STATE $G(c_i)\gets\{k:\ u^{(G)}_k<\gamma_g\}$;\quad
         $R(c_i)\gets\{k\in\overline{G}:\ u^{(YR)}_k<\tfrac{\gamma_r}{1-\gamma_g}\}$
  \STATE $s^G_i \leftarrow \mathbf{1}[x_i \in G(c_i)]$;\quad
         $s^R_i \leftarrow \mathbf{1}[x_i \in R(c_i)]$
\ENDFOR
\STATE $L \leftarrow N$
\STATE $S_G \leftarrow \sum_{i=1}^{N} s^G_i$;\quad
       $S_R \leftarrow \sum_{i=1}^{N} s^R_i$
\STATE $\hat p_G \leftarrow S_G/L$;\quad
       $\hat p_R \leftarrow S_R/L$
\STATE $z_G \leftarrow \dfrac{\hat p_G-\gamma_g}{\sqrt{\gamma_g(1-\gamma_g)/L}}$
\STATE $z_R \leftarrow \dfrac{\gamma_r-\hat p_R}{\sqrt{\gamma_r(1-\gamma_r)/L}}$
\STATE $p_G \leftarrow 1-\Phi(z_G)$;\quad
       $p_R \leftarrow 1-\Phi(z_R)$
\STATE $\text{score} \leftarrow -2\!\big(\lambda_f \ln p_G + (1-\lambda_f)\ln p_R\big)$
\STATE $\tau_\alpha \leftarrow \chi^2_{4}(1-\alpha)$ \hfill{\small }
\STATE \textbf{return} \textsc{Watermarked} if $\text{score} \ge \tau_\alpha$; else \textsc{Clean}
\end{algorithmic}
\end{algorithm}

This assumption brings three practical benefits.\textbf{(i) Power under realism:} by accounting for stepwise variability, the test avoids optimistic or pessimistic tails that would otherwise inflate or deflate $p$-values, yielding higher power at the same FPR. \textbf{(ii) Seamless with our pipeline:} the Poisson–binomial view naturally accommodates sliding windows, entropy-gated bias, special-token masks, and dual evidence (Green enrichment + Red depletion), without extra heuristics.

\paragraph*{Fisher combination.}
Green enrichment and Red depletion provide complementary evidence.
We combine their one-sided $p$-values via Fisher’s score
\begin{equation}
S_\lambda \;=\; -2\!\bigl(\lambda \ln p_G + (1-\lambda)\ln p_R\bigr),
\end{equation}
which increases when either source is more significant and still rewards cases where both are moderate.
Under weak dependence, $S_\lambda$ is well-approximated by $\chi^2_{(4)}$, yielding a closed-form, globally calibrated threshold after multi-window correction.

\section{Experiments}
\subsection{Experimental Setup}

In this section, we further validate the effectiveness of our method based on the \textbf{LLaMA 2 7B} model~\cite{b20}. Specifically, we use \texttt{Llama-2-7b-chat-hf} for generating watermarked text, while \texttt{Llama-2-7b-hf} is used to compute perplexity. We evaluate the algorithm from three perspectives: detection accuracy (True Positive Rate), false positive rate (False Positive Rate), and text quality.

\paragraph*{Implementation and Environment.}
We implement the watermark on top of \textbf{PyTorch} and the \textbf{Hugging Face Transformers} backend~\cite{b25}. On the generation side, we rely on the \texttt{generate} API and a custom \texttt{LogitsProcessor} to warp model logits; tokenization uses \texttt{AutoTokenizer}, and models are loaded via \texttt{AutoModelForCausalLM}. Detection replays the same partitioner and computes sliding-window statistics in Python with \textbf{NumPy}, and we produce figures with \textbf{Matplotlib}. All runs use the \textbf{torch} random number generator for seeding.

\paragraph*{Datasets and Prompts.}
The \textbf{Alpaca} dataset~\cite{b21} provides 52K high-quality instruction–response pairs generated through the Self-Instruct framework. It covers a broad range of open-domain tasks such as reasoning, summarization, and classification, and demonstrates strong linguistic diversity and instruction fidelity. Because of its diversity, realism, and accessibility, we select Alpaca to evaluate watermark embedding and detection under realistic instruction-following scenarios. For each compared method, we generate 600 samples, each with approximately 250 tokens.

To assess the false positive rate, we further construct a mixed dataset by combining samples from four public corpora: the \textbf{WikiText} language modeling dataset extracted from high-quality Wikipedia articles~\cite{b22}, the \textbf{IMDB Movie Reviews} dataset for sentiment classification~\cite{b23}, the \textbf{AG News} corpus covering four categories (\textit{World}, \textit{Sports}, \textit{Business}, and \textit{Science/Technology})~\cite{b24}, and the \textbf{Yelp Polarity} dataset commonly used for sentiment analysis~\cite{b24}. We randomly sample 600 texts from these corpora, each around 250 tokens, to form a balanced evaluation set. This mixed dataset exhibits diverse writing styles, varied lengths and structures, multiple task types, reduced corpus bias, and strong representativeness, enabling a more comprehensive and reliable assessment of false positive rates across heterogeneous linguistic distributions.
\paragraph*{Compared methods}
We compare our method with several representative watermarking approaches, namely 
KGW~\cite{b7}, KGW2~\cite{b16}, OpenAI~\cite{b19}, and Marylandz~\cite{b19}.

\paragraph{True Positive Rate}
We first evaluate the detection success rate. 
For each method, we generate watermarked texts (completions) using its respective watermarking algorithm 
and subsequently apply the corresponding detection procedure. 
Since all generated completions are produced by watermark-enabled models, 
each text should, in principle, contain an embedded watermark. 
Therefore, the proportion of correctly detected watermarked texts represents 
the \textit{true positive rate (TPR)} for each method.

\begin{table}[h]
\centering
\caption{TPR Comparison Among Watermarking Methods}
\label{tab:tpr}
\setlength{\tabcolsep}{3pt} 
\renewcommand{\arraystretch}{0.95} 
\begin{tabular*}{\columnwidth}{@{\extracolsep{\fill}}lccccc}
\toprule
Metric & KGW & KGW2 & OpenAI & Marylandz & Ours \\
\midrule
True positives & 233/600 & 96/600 & 117/600 & 121/600 & \textbf{370/600} \\
TPR (\%) & 38.83 & 16.00 & 19.50 & 20.17 & \textbf{61.67} \\
\bottomrule
\end{tabular*}
\end{table}

As shown in Table~\ref{tab:tpr}, our method achieves the highest detection success rate, 
outperforming prior watermarking baselines.

\paragraph{False Positive Rate}
We evaluate the false positive rate on a mixed human-written corpus that contains no watermarks. 
Since all texts are negatives by construction, the expected FPR is essentially zero; hence, 
lower values indicate better performance. For each method, we run its detector on 600 samples 
and report the proportion that is (incorrectly) flagged as watermarked.


\begin{table}[h]
\centering
\caption{FPR Comparison Among Watermarking Methods}
\label{tab:fpr}
\setlength{\tabcolsep}{3pt}
\renewcommand{\arraystretch}{0.95}
\begin{tabular*}{\columnwidth}{@{\extracolsep{\fill}}lccccc}
\toprule
Metric & KGW & KGW2 & OpenAI & Marylandz & Ours \\
\midrule
False positives & 12/600 & 12/600 & 15/600 & 9/600 & \textbf{3/600} \\
FPR (\%) & 2.00 & 2.00 & 2.50 & 1.50 & \textbf{0.50} \\
\bottomrule
\end{tabular*}
\end{table}

Results in Table~\ref{tab:fpr} indicate that our method yields the lowest false positive rate (FPR) 
among all compared approaches, effectively reducing false detections on clean, non-watermarked texts 
while maintaining reliable discrimination capability.

\paragraph{Text Quality}
We further evaluate the quality of generated texts depending on \textit{perplexity}. 

\begin{figure}[t]
  \centering
  \includegraphics[width=\columnwidth]{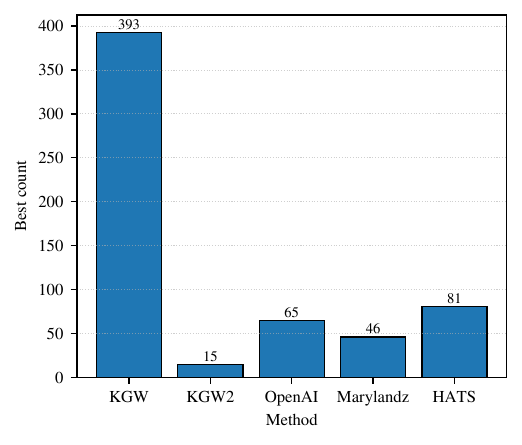} 
  \caption{Perplexity-based best-count (number of texts) for KGW, KGW2, OpenAI, Marylandz, and HATS. Lower perplexity indicates better alignment with the model’s language distribution.}
  \label{fig:best_counts}
\end{figure}

Perplexity measures the fluency and naturalness of generated text, reflecting how confidently a language model predicts the next token in a sequence.
It is defined as:
\begin{equation}
\mathrm{PPL} = \exp\!\left(-\frac{1}{N}\sum_{i=1}^{N}\log p(x_i)\right),
\end{equation}
where $N$ denotes the total number of tokens and $p(x_i)$ is the model-assigned probability for the $i$-th token.
A lower perplexity indicates that the text better aligns with the language distribution captured by the model, 
therefore corresponding to higher fluency and coherence.
This metric is widely adopted in natural language generation tasks as an intrinsic measure of linguistic plausibility.
We compute perplexity using the Llama-2-7b-hf model to ensure consistency with large-scale generative benchmarks.
Each generation file is ordered such that every line corresponds to the same prompt across methods. For each line, we compute the completion’s perplexity, compare the perplexities of all methods on that line, and record the method with the lowest value. We then count how many times each method wins and visualize the results as a bar chart.
Results in Fig.~\ref{fig:best_counts} show that although \textit{HATS} does not achieve the lowest perplexity, it ranks second in the comparison, outperforming the remaining methods.

Overall, although our proposed \textbf{HATS} watermarking method does not yet match the semantic quality achieved by KGW, it attains the best performance among all compared approaches in terms of detection success rate and false positive rate, demonstrating its strong detectability and robustness.

For future work, we plan to further enhance the semantic naturalness and contextual coherence of the generated text, while establishing a comprehensive, multi-metric evaluation framework to assess text quality from the perspectives of fluency, readability, and semantic fidelity. In addition, we will strengthen the reliability and robustness of our method to ensure stable detection performance across different models, perturbation types, and adversarial scenarios.

\section*{Acknowledgment}
This work was supported by the National Natural Science Foundation of China. 
The authors thank the faculty members and students of the Neuron Insight Lab 
for their helpful assistance and insightful discussions during this research. 
The authors also thank Associate Researcher Xiaolei Liu for valuable guidance and constructive feedback throughout the project.

\end{document}